\documentclass[sigconf,authordraft,review=false]{acmart}
\usepackage{algorithm}
\usepackage{algorithmic}
\usepackage{multirow}
\usepackage{enumitem}
\usepackage{dsfont}

\AtBeginDocument{%
  }

\setcopyright{acmlicensed}
\copyrightyear{2018}
\acmYear{2018}
\acmDOI{XXXXXXX.XXXXXXX}
\acmConference[Conference acronym 'XX]{Make sure to enter the correct
  conference title from your rights confirmation emai}{June 03--05,
  2018}{Woodstock, NY}
\acmISBN{978-1-4503-XXXX-X/18/06}
\newcommand{\method}{KEDRec-LM}

\begin{document}

\title{\method: A \underline{K}nowledge-distilled \underline{E}xplainable \underline{D}rug \underline{Rec}ommendation Large \underline{L}anguage \underline{M}odel}

\author{Kai Zhang}
\email{kzhang8@wpi.edu}
\affiliation{%
  \institution{Worcester Polytechnic Institute}
  \city{Worcester}
  \country{USA}
}

\author{Rui Zhu}
\affiliation{%
  \institution{Yale University}
  \city{New Heaven}
  \country{USA}
}
\email{rui.zhu.rz399@yale.edu}

\author{Shutian Ma}
\affiliation{%
  \institution{}
  \city{Bloomington}
  \country{USA}
}
\email{mashutian0608@hotmail.com}

\author{Jingwei Xiong}
\affiliation{%
 \institution{University of California, Davis}
 \city{Davis}
 \country{USA}}
\email{jwxxiong@ucdavis.edu}

\author{Yejin Kim}
\email{Yejin.Kim@uth.tmc.edu}
\affiliation{%
  \institution{The University of Texas Health Science Center at Houston}
  \city{Huston}
  \country{USA}}

\author{Fabricio Murai}
\email{fmurai@wpi.edu}
\affiliation{%
  \institution{Worcester Polytechnic Institute}
  \city{Worcester}
  \country{USA}
}

\author{Xiaozhong Liu}
\email{xliu14@wpi.edu}
\affiliation{%
  \institution{Worcester Polytechnic Institute}
  \city{Worcester}
  \country{USA}
}

\begin{abstract}
Drug discovery is a critical task in biomedical natural language processing (NLP), yet explainable drug discovery remains underexplored. Meanwhile, large language models (LLMs) have shown remarkable abilities in natural language understanding and generation. Leveraging LLMs for explainable drug discovery has the potential to improve downstream tasks and real-world applications. In this study, we utilize open-source drug knowledge graphs, clinical trial data, and PubMed publications to construct a comprehensive dataset for the explainable drug discovery task, named \textbf{expRxRec}. Furthermore, we introduce \textbf{\method{}}, an instruction-tuned LLM which distills knowledge from rich medical knowledge corpus for drug recommendation and rationale generation. To encourage further research in this area, we will publicly release\footnote{A copy is attached with this submission} both the dataset and \method{}.
\end{abstract}


\maketitle

\section{Introduction}
The complexity of drug discovery lies in understanding the intricate relationships between drugs and diseases, making the identification of potential therapeutic uses a challenging and resource-intensive endeavor. In recent years, the availability of large-scale biomedical knowledge graphs~\cite{pubmed_kg_2024, kg_ctg_2024, citation_sum_2023}, such as the Drug Repurposing Knowledge Graph (DRKG)~\cite{drkg}, has enabled significant advances in this field by linking vast amounts of biomedical entities and relationships. These structured databases capture a wealth of information across drug interactions, disease associations, and biological pathways. Yet, fully exploiting this information for drug discovery, and more specifically drug repurposing, requires efficient methods to extract meaningful insights that can guide therapeutic reasoning.

With the rapid growth of biomedical literature, particularly in repositories like PubMed, there is an unprecedented opportunity to tap into this body of knowledge to inform drug-disease relationships. However, manually curating, understanding, and drawing conclusions from this literature is impractical due to its sheer volume and the specificity of each study. Traditional knowledge graphs provide a static representation of these connections, but their utility is limited when it comes to reasoning about complex therapeutic mechanisms or the nuanced interplay between drug efficacy and disease pathology. This gap highlights the need for automated approaches that can contextualize and synthesize existing literature to support and enhance the process of drug design and discovery.

The challenges in leveraging biomedical literature for drug discovery are multifaceted. Firstly, the process of identifying and validating drug-disease pairs from a vast knowledge graph is inherently complex, given the heterogeneity of relationships and the potential noise in the data~\cite{Peng2017, Wang2020, Zhu2019}. Secondly, even when relevant pairs are identified, understanding their context and implications requires sifting through large volumes of literature, which can vary greatly in quality, relevance, and specificity~\cite{haddaway2020eight,unfoldai2023rag}. Standard retrieval-based systems often fall short in providing insightful reasoning over the retrieved content. Moreover, transforming this wealth of information into a format that can be easily used for computational models in drug design remains an unmet challenge. In essence, there is a critical need for a systematic, automated approach to extract, understand, and reason over biomedical literature in a way that is aligned with the complexities of drug-disease associations~\cite{huang2024computational,bblgat2022, gendrin2023investigating}.

To address these challenges, we propose a novel framework that integrates knowledge graph extraction, literature mining, and language model-based reasoning. Our approach begins by extracting a subset of drug-disease pairs from the DRKG, thereby forming a focused sub-knowledge graph tailored for drug discovery. For each pair in this graph, we employ a Retrieval-Augmented Generation (RAG)~\cite{lewis2020retrieval} technique which searches for relevant literature content in PubMed and Clinical Trials, leveraging the wealth of biomedical literature to contextualize each drug-disease relationship. To facilitate reasoning over these pairs, we design a set of domain-specific questions that are crucial for understanding the therapeutic potential of each relationship. An instructional template is then used to fit the pairs as well as retrieved background information for a teacher model generating responses to these questions. These responses provide rich, contextually informed insights into each pair, transforming unstructured text into structured, reasoned answers.

Building on this, we use these generated responses to train a LLaMA model~\cite{touvron2023llama2} as the local model for learning those distilled knowledge. This model is designed to understand and reason over drug-disease relationships, providing a tool that not only synthesizes biomedical knowledge but also supports hypothesis generation and decision-making in drug design. By leveraging both structured knowledge graphs and unstructured literature, our approach enables a deeper understanding of drug-disease associations, facilitating the discovery of novel therapeutic opportunities.

This paper makes the following contributions:
\begin{itemize}[left=0pt .. \parindent]
\item \textbf{RAG-Based Literature Reasoning} We design domain-specific questions and utilize a GPT-based Retrieval-Augmented Generation (RAG) pipeline to extract and reason over biomedical literature from PubMed for each drug-disease pair.

\item \textbf{Domain-Specific local LLM} We use the structured responses generated by RAG and a powerful teacher model to train a specialized LLaMA model, tailored for drug design reasoning. This model captures nuanced drug-disease relationships and supports inference-making in therapeutic research.

\item \textbf{Experimental Resources} We demonstrate the effectiveness of our approach in explainable drug recommendation, open-sourcing a new dataset and a local model to the community. 
\end{itemize}
Our work represents a step forward in the integration of knowledge graphs, literature mining, and language model-based reasoning for explainable drug recommendation.

\section{Background}

\textbf{Large Language Models and Retrieval-Augmented Generation.}
Large Language Models (LLMs) are deep neural networks trained on massive corpora of text data to generate and understand human-like language. By learning statistical patterns in text, LLMs, such as GPT and BioGPT, are capable of tasks like text generation, summarization, and reasoning. Given an input sequence $q$, an LLM generates a probability distribution over possible outputs $y$, modeled as:
\[
p(y|q) = \prod_{t=1}^T p(y_t|q, y_{<t}),
\]
where $T$ is the output sequence length, and $y_{<t}$ represents the tokens generated up to time $t$.

However, standard LLMs are limited by the knowledge encoded during training and may struggle with domain-specific tasks such as biomedical reasoning. Retrieval-Augmented Generation (RAG) addresses this limitation by combining retrieval-based and generation-based methods, enabling LLMs to incorporate external knowledge in real-time. The workflow of RAG can be broken into two stages:
\begin{enumerate}
    \item \textbf{Retrieval}: For a given query $q$, relevant documents are retrieved from an external corpus $\mathcal{D}$. Each document $d_i \in \mathcal{D}$ is scored based on the similarity between the query embedding $\mathbf{e}(q)$ and document embedding $\mathbf{e}(d_i)$. Cosine similarity is used for scoring:
    \[
    \text{sim}(q, d_i) = \frac{\mathbf{e}(q) \cdot \mathbf{e}(d_i)}{\|\mathbf{e}(q)\| \|\mathbf{e}(d_i)\|}.
    \]
    The top-$k$ documents with the highest scores are selected as the retrieval set $\mathcal{R}$.
    
    \item \textbf{Generation}: The query $q$ and the retrieved context $\mathcal{R}$ are concatenated and fed into the LLM. The LLM generates an output $y$ conditioned on both the query and the retrieved documents:
    \[
    p(y|q, \mathcal{R}) = \prod_{t=1}^T p(y_t|q, \mathcal{R}, y_{<t}).
    \]
\end{enumerate}

This approach improves the factual accuracy and interpretability of LLM outputs by grounding predictions in external knowledge. By integrating the retrieval capability with the language generation power of LLMs, RAG enhances reasoning and ensures that predictions are backed by relevant evidence.\\

\noindent \textbf{Knowledge Distillation.}
Knowledge Distillation is a training paradigm where a smaller, student model $f_\theta$ learns from a larger, more powerful teacher model $f_T$. The objective is to transfer the teacher model's knowledge while optimizing the student model's efficiency and performance.  

In this study, the teacher model supervises two tasks: drug selection and rationale generation. The loss function for knowledge distillation consists of:
\begin{enumerate}
    \item \textbf{Selection Loss}: The student model predicts a probability $p(c^* | d, c_1, c_2)$ for selecting the correct drug candidate $c^*$ from a given set $(d, c_1, c_2)$, matching the teacher's output. This loss is defined as:
    \[
    \mathcal{L}_{\text{select}} = -\log p(c^* | d, c_1, c_2).
    \]

    \item \textbf{Rationale Generation Loss}: The student generates a rationale $r$ explaining its selection, aligning with the teacher-provided rationale $r_T$. The loss is defined as:
    \[
    \mathcal{L}_{\text{rationale}} = \|r - r_T\|^2,
    \]
    where $r$ and $r_T$ are embeddings of the student and teacher rationales, respectively.
\end{enumerate}

The overall training objective for the student model is:
\[
\mathcal{L}(\theta) = \mathcal{L}_{\text{select}} + \lambda \mathcal{L}_{\text{rationale}},
\]
where $\lambda$ is a hyperparameter balancing the importance of the two components.

Through knowledge distillation, the student model learns to replicate the teacher's performance in drug selection and rationale generation while improving computational efficiency. Paired with RAG, this allows the system to provide accurate and interpretable drug recommendations grounded in biomedical knowledge.

\section{Methodology}
\begin{figure}
    \centering
    \includegraphics[width=\linewidth]{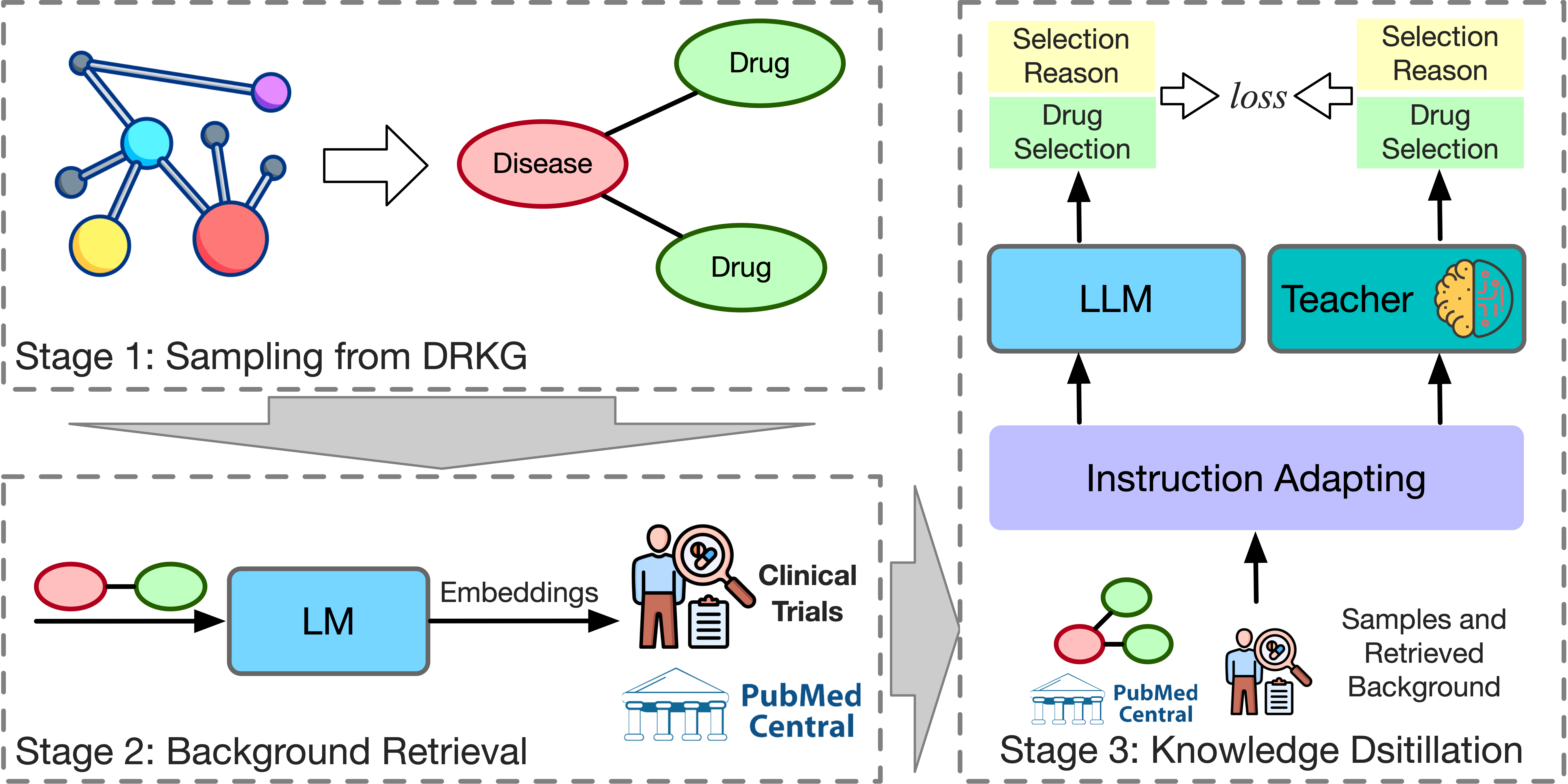}
    \caption{Overview of KRxELM. Stage 1 samples a disease-drug set which consists of a disease and two drug candidates from the open-sourced rich DRKG; Stage 2 leverages a Language model to embed the drug-disease pair from the sampled disease-drug set, the embeddings are further used for obtaining background information from Clinical Trials and PubMed Central corpora; Stage 3 fits the disease-drug set and the retrieved background information into an instructional prompt template as input to a large language model (student) and a teacher model. Instruction tuning was used to enable the LLM to learn from the teacher model.}
    \label{fig:methodology}
\end{figure}

\textbf{Task Formation} In this study, we address the problem of drug discovery by leveraging large language models (LLMs) to select optimal drug candidates and provide explainable reasoning. Given a disease-drug set $S = (d, c_1, c_2)$, where $d$ represents the disease and $c_1, c_2$ are the two drug candidates, the task is to train an LLM $f_\theta$ to achieve two objectives: (1) to select the most suitable drug candidate $c^* \in \{c_1, c_2\}$ for the disease $d$, and (2) to generate an explainable rationale $r$ for the selection.\\ 

\noindent \textbf{\method{}}. We introduce a new instruction-tuned LLM which distills knowledge from rich medical knowledge corpus for drug recommendation and rationale generation. \method{} consists of 3 stages shown in Fig.~\ref{fig:methodology}, which we describe next.



\subsection{Stage 1: Sampling from DRKG}
Our disease-drug set $S$ is sampled from the open-sourced rich knowledge graph DRKG\footnote{https://github.com/gnn4dr/DRKG} (Drug Repurposing Knowledge Graph) which is a comprehensive biological knowledge graph relating genes, compounds, diseases, biological processes, side effects and symptoms. The sampling process involves selecting one relevant drug candidate $c_{\text{rel}}$ and one irrelevant drug candidate $c_{\text{irr}}$ for a given disease $d$. To ensure the selection is challenging, the irrelevant candidate $c_{\text{irr}}$ is chosen such that it is similar to $c_{\text{rel}}$ based on their embeddings, which are learned using a graph neural network (GNN)-based model. Let $\mathbf{h}_c$ denote the embedding of a drug candidate $c$ learned by the GNN. The similarity between two drug candidates is computed as:
\begin{equation}
\text{sim}(c_1, c_2) = \frac{\mathbf{h}_{c_1} \cdot \mathbf{h}_{c_2}}{\|\mathbf{h}_{c_1}\| \|\mathbf{h}_{c_2}\|},
\end{equation}
where $\cdot$ denotes the dot product and $\|\cdot\|$ represents the Euclidean norm.

For a given disease $d$, the GNN-based model identifies a set of candidate drugs $C_d$ from DRKG. The relevant candidate $c_{\text{rel}}$ is selected as:
\begin{equation}
c_{\text{rel}} = \arg\max_{c \in C_d} \text{rel}(d, c),
\end{equation}
where $\text{rel}(d, c)$ is a relevance score derived from DRKG edges connecting $d$ and $c$. The irrelevant candidate $c_{\text{irr}}$ is then sampled as:
\begin{equation}
c_{\text{irr}} = \arg\max_{c \in C_d \setminus \{c_{\text{rel}}\}} \text{sim}(c, c_{\text{rel}}),
\end{equation}
ensuring that $c_{\text{irr}}$ is similar to $c_{\text{rel}}$ but not relevant to $d$.

\subsection{Stage 2: Background Retrieval}
The sampled disease-drug set $S$, consisting of pairs $(d, c_i)$ for $i \in \{1, 2\}$, is then used for retrieving background information from two rich corpora: Clinical Trials\footnote{https://clinicaltrials.gov/} and PubMed Central\footnote{https://ftp.ncbi.nlm.nih.gov/pub/pmc/oa\_bulk/oa\_comm/xml/}. A language model $g_\phi$ is employed as an embedding projection mechanism to facilitate the retrieval process. For each pair $(d, c_i)$, the background information $I_i$ is retrieved by computing the relevance between the pair's embedding $\mathbf{e}(d, c_i)$ and the embeddings of documents in the corpora. Let $\mathbf{e}_j$ denote the embedding of the $j$-th document in the corpus. The relevance score is computed as:
\begin{equation}
\text{rel}(\mathbf{e}(d, c_i), \mathbf{e}_j) = \frac{\mathbf{e}(d, c_i) \cdot \mathbf{e}_j}{\|\mathbf{e}(d, c_i)\| \|\mathbf{e}_j\|}.
\end{equation}
The top-$k$ documents with the highest relevance scores are selected as the background information $I_i$. This retrieved information serves as an instructional corpus for fine-tuning the LLM, ensuring that the model incorporates domain-specific knowledge. 

The learning process is guided by a teacher model $f_T$, which provides supervisory signals for both candidate selection and rationale generation. The optimization objective is to minimize the following loss function:
\begin{equation}
\mathcal{L}(\theta) = \mathcal{L}_{\text{select}}(c^*, f_T) + \mathcal{L}_{\text{rationale}}(r, f_T),
\end{equation}
where $\mathcal{L}_{\text{select}}$ quantifies the alignment of $f_\theta$'s candidate selection with the teacher model's predictions, and $\mathcal{L}_{\text{rationale}}$ measures the quality of the generated rationale compared to teacher-provided explanations.

Through this approach, we aim to create a robust and interpretable model for aiding drug discovery by synthesizing insights from clinical and biomedical literature.

\subsection{Stage 3: Knowledge Distillation}
The sampled disease-drug set $S$ and the retrieved background information $\{I_1, I_2\}$ are incorporated into an instructional template to construct input sequences for training the LLM. Each input consists of the disease $d$, the drug candidates $c_1, c_2$, and the corresponding background information $I_1, I_2$. The output of the LLM $f_\theta$ includes the selected drug candidate $c^*$ and a generated rationale $r$ explaining the decision.

A powerful teacher model $f_T$ is used to supervise the training process. The LLM is enforced to learn from the teacher model using instruction fine-tuning. The training objective is designed to minimize two types of loss:
\begin{itemize}
    \item \textbf{Reason generation loss:} This loss, denoted as $\mathcal{L}_{\text{rationale}}$, measures the discrepancy between the rationale $r$ generated by the LLM and the rationale $r_T$ provided by the teacher model:
    \begin{equation}
    \mathcal{L}_{\text{rationale}} = \|r - r_T\|^2.
    \end{equation}
    \item \textbf{Drug selection loss:} This loss, denoted as $\mathcal{L}_{\text{select}}$, captures the difference between the drug selection $c^*$ made by the LLM and the ground truth drug candidate $c_{\text{rel}}$ derived from the sampling process:
    \begin{equation}
    \mathcal{L}_{\text{select}} = \mathds{1}[c^* \neq c_{\text{rel}}],
    \end{equation}
    where $\mathds{1}[\cdot]$ is an indicator function that evaluates to 1 if the condition is true and 0 otherwise.
\end{itemize}
The total loss for training the LLM is given by:
\begin{equation}
\mathcal{L}(\theta) = \mathcal{L}_{\text{select}} + \lambda \mathcal{L}_{\text{rationale}}\,,
\end{equation}
where $\lambda$ is a weighting factor to balance the two losses.

\section{Experiments}
\subsection{Dataset \& Experimental Setup}
\textbf{expRxRec}. To predict the effects of drug compounds for specific diseases, we design a data pipeline that integrates structured knowledge from an existing biomedical knowledge graph and unstructured information from medical corpus.\par
\textbf{Disease-Drug Compound Pairs Sampling from DRKG}. For each disease, we select two types of drug compounds: one with a positive effect and another with a negative effect or a comparatively less positive effect based on the connectivity within the graph. This method enables the model to distinguish the drug candidates in terms of a given disease. \par
\textbf{Enriching the Dataset with Background Retrieval} To enhance the information of disease-drug relationships, we extract relevant information from PubMed Central and Clinical Trials using RAG from the open-sourced raw data. A cleaning process is further applied to ensure data relevance and quality:
\begin{enumerate}
    \item Articles with empty titles or abstracts are removed.
    \item Articles are retained only if their content contains at least one sampled drug name.
\end{enumerate}
Following this process, 1,905,387 articles are retained. We then use Apache Lucene\footnote{https://lucene.apache.org/}, a high-performance text search library, to index the core content of the articles. To ensure that queries could focus on the most relevant biomedical information, a two-step retrieval process is conducted:
\begin{enumerate}
    \item For each disease-drug compound pair, we construct a query comprising the disease name and the compound name. The Lucene \texttt{TopDocs} is utilized to retrieve the top-\(k\) most relevant text chunks, where \(k\) is set to 80 in this study.
    \item To further refine the retrieved information, the top-\(k\) text chunks are embedded using OpenAI’s vector embedding models, which capture semantic representations of the text. A similarity filter is applied to retain chunks that are most relevant to the disease-drug compound pair.
\end{enumerate}
The final dataset thus contains: a disease name, a drug compound name, the relationship label (positive or negative effect), and a curated set of PubMed and Clinical Trials chunks, providing context and supporting evidence. This enriched dataset forms the input for training a robust large language model (LLM) capable of reasoning over biomedical data, enabling it to predict the effects of drug compounds for given diseases.\\
\textbf{MIMIC-III} is a publicly available resource of de-identified health data from over 58,000 critical care admissions at Beth Israel Deaconess Medical Center~\citep{johnson2016mimic}. It combines structured data (e.g., lab results, prescriptions) and unstructured clinical notes, offering a rich source for research in drug discovery and patient-specific treatment analysis while maintaining HIPAA compliance.

\subsection{Results on Drug Selection}
Table 1 presents the F1 scores for various models on the drug selection task, evaluated across two datasets: expRxRec and MIMIC-III. Each model's performance is assessed under four configurations: vanilla (baseline without background information), with Clinical Trials (CT), with PubMed Central (PMC), and with both CT and PMC.

Across all models, incorporating background information (CT, PMC, or both) consistently improves performance over the vanilla version. For example, GNN achieves an F1 score improvement from 76.66 to 84.91 on expRxRec and from 57.05 to 64.01 on MIMIC-III when enriched with both CT and PMC. Similarly, \method{} demonstrates the best overall performance, achieving the highest F1 scores of 88.05 (expRxRec) and 69.19 (MIMIC-III) with both CT and PMC.

Notably, while the inclusion of either CT or PMC individually enhances the results, combining both sources yields the largest improvements. This trend highlights the complementary nature of the two sources in enriching the input data for better drug selection predictions. Furthermore, the performance gap between expRxRec and MIMIC-III suggests that the structured information and scale of expRxRec may facilitate better model generalization compared to the more diverse and sparse data in MIMIC-III.

In summary, these results emphasize the importance of leveraging external background knowledge for improving drug selection accuracy. Among the tested models, \method{} exhibits superior performance, showcasing its efficacy in integrating background information for this task.
\begin{table}[ht]
    \small
    \centering
    \begin{tabular}{cccc}
        \hline
         Model & Type & expRxRec & MIMIC-III \\
          \hline
         \multirow{4}{*}{GNN\citep{gaudelet2021utilizing}} & \textit{vanilla} & 76.66 & 57.05  \\
         & \textit{w CT} & 81.37 & 61.95  \\
         & \textit{w PMC} & 81.94 & 61.55  \\
         & \textit{w both} & 84.91 & 64.01  \\
         \hline
         \multirow{4}{*}{SafeDrug\citep{yang2021safedrug}} & \textit{vanilla} & 76.21 & 64.85 \\
         & \textit{w CT} & 82.09 & 66.91 \\
         & \textit{w PMC} & 81.20 & 66.81 \\
         & \textit{w both} & 83.20 & 67.77 \\
         \hline
         \multirow{4}{*}{4SDrug\citep{tan20224sdrug}} & \textit{vanilla} & 76.71 & 61.96 \\
         & \textit{w CT} & 82.12 & 64.01 \\
         & \textit{w PMC} & 82.27 & 64.13 \\
         & \textit{w both} & 83.71 & 66.29 \\
         \hline
         \multirow{4}{*}{\method{}} & \textit{vanilla} & 80.11 & 65.10 \\
         & \textit{w CT} & 87.44 & 68.43 \\
         & \textit{w PMC} & 87.61 & 68.71 \\
         & \textit{w both} & \textbf{88.05} & \textbf{69.19} \\
         \hline
    \end{tabular}
    \caption{The results on drug selection with F1 score equipped with different retrieved background information.}
    \vspace{-10pt}
    \label{tab:drug_selection}
\end{table}

\subsection{Results on Reason Generation}
\begin{table}[ht]
    \small
    \centering
    \begin{tabular}{ccccc}
        \hline
         Model & Type & ROUGE-1 & ROUGE-2 & ROUGE-L \\
          \hline
         \multirow{4}{*}{Pointer-Generator\citep{see2017get}} & \textit{vanilla} & 15.11 & 15.61 & 12.29 \\
         & \textit{w CT} & 17.03 & 17.61 & 15.29 \\
         & \textit{w PMC} & 17.91 & 17.94 & 15.02 \\
         & \textit{w both} & 18.33 & 18.67 & 16.91 \\
         \hline
         \multirow{4}{*}{BioGPT\citep{luo2022biogpt}} & \textit{vanilla} & 22.19 & 22.42 & 20.16 \\
         & \textit{w CT} & 26.79 & 26.99 & 25.01 \\
         & \textit{w PMC} & 25.87 & 25.89 & 23.97 \\
         & \textit{w both} & 27.19 & 27.58 & 25.81 \\
         \hline
         \multirow{4}{*}{\method{}} & \textit{vanilla} & 28.90 & 28.14 & 26.43 \\
         & \textit{w CT} & 33.03 & 33.78 & 31.74 \\
         & \textit{w PMC} & 33.89 & 33.95 & 31.90 \\
         & \textit{w both} & \textbf{35.11} & \textbf{35.17} & \textbf{34.03} \\
         \hline
    \end{tabular}
    \caption{The results on recommendation reason generation with ROUGE scores equipped on expRxRec.}
    \vspace{-8pt}
    \label{tab:reason_generation}
\end{table}

Table 2 presents the ROUGE-1, ROUGE-2, and ROUGE-L scores for recommendation reason generation using Pointer-Generator, BioGPT, and \method{} under the same four configurations as in the drug selection task.

The inclusion of background knowledge, whether from Clinical Trials or PubMed Central, consistently improves the performance of all models compared to the vanilla baseline. The Pointer-Generator model, which serves as a sequence-to-sequence baseline, shows modest gains with background enrichment, improving its ROUGE-1 score from 15.11 (vanilla) to 18.33 (w both). However, its overall performance remains limited due to its lack of domain-specific pretraining. BioGPT, a biomedical pre-trained language model, performs better than Pointer-Generator across all configurations. In the vanilla setting, it achieves a ROUGE-1 score of 22.19, which is significantly higher than Pointer-Generator, indicating the benefit of pretraining on biomedical text. However, its improvement with background enrichment is less pronounced, with a ROUGE-1 score of 27.19 when both CT and PMC are used. This suggests that BioGPT leverages its internal knowledge more effectively but does not fully exploit external background information. \method{} achieves the best performance across all metrics and configurations, demonstrating its superior ability to generate coherent and contextually rich explanations. In the vanilla setting, \method{} achieves a ROUGE-1 score of 28.90, outperforming both Pointer-Generator and BioGPT by a significant margin. When enriched with both CT and PMC, \method{} reaches the highest scores: 35.11 (ROUGE-1), 35.17 (ROUGE-2), and 34.03 (ROUGE-L). These results underscore the model's capability to effectively integrate and distill information from diverse knowledge sources, enabling it to generate high-quality, explainable recommendations.

The combination of Clinical Trials and PubMed Central provides the most comprehensive improvement, highlighting the complementary nature of these knowledge sources. Clinical Trials offer structured and detailed information on drug efficacy, while PubMed Central provides broader contextual insights. For example, \method{}'s performance improves significantly from 33.03 (w CT) and 33.89 (w PMC) to 35.11 (w both), demonstrating the benefit of combining these sources.

Overall, these findings suggest that integrating external knowledge, particularly from multiple complementary sources, is crucial for enhancing the explainability and quality of recommendation generation. Furthermore, \method{}'s superior performance highlights the importance of designing models specifically tailored for biomedical reasoning tasks, enabling them to effectively synthesize and reason over complex biomedical information.

\section{Related Work}
The integration of large language models (LLMs) and knowledge graphs has significantly advanced drug discovery and repurposing. LLMs, with their capacity to process and reason over unstructured biomedical data, have shown promise in uncovering target-disease linkages and facilitating clinical trial optimization. ~\citet{zheng2024large} provide an overview of LLM applications in drug development, while \citet{gangwal2024generative} demonstrate the ability of generative transformers to design novel drug molecules using extensive biomedical datasets.

Knowledge graphs have also played a crucial role in drug repurposing. The Drug Repurposing Knowledge Graph (DRKG)~\citep{drkg}, which integrates data from multiple sources, including DrugBank~\citep{drugbank} and Hetionet~\citep{himmelstein2017systematic}, enables computational methods to identify novel drug-disease associations. Techniques such as graph neural networks (GNNs) have further improved prediction accuracy~\citep{doshi2022computational}. Additionally, hybrid approaches that combine LLMs with knowledge graphs, such as those proposed by ~\citet{fei2021enriching}, enhance reasoning over structured and unstructured biomedical data.

While these approaches emphasize either LLMs or knowledge graphs independently, our work uniquely combines these technologies with retrieval-augmented generation (RAG)~\citep{lewis2020retrieval} to incorporate structured and unstructured background information. Unlike prior works, we focus on building a framework that enables explainable drug recommendation by integrating knowledge distillation techniques to improve model interpretability and decision-making. By leveraging background retrieval from PubMed and Clinical Trials, our approach bridges the gap between LLM reasoning and biomedical knowledge, offering a robust solution for distinguishing drug candidates with supporting evidence.

\vspace{-8pt}
\section{Conclusion}
In this study, we introduce \method{}, a novel framework designed to advance drug discovery by integrating structured biomedical knowledge and unstructured textual data. By leveraging the Drug Repurposing Knowledge Graph (DRKG) and enriching it with relevant contextual information from PubMed Central and Clinical Trials, we curated a comprehensive dataset that enhances the accuracy of drug recommendation and the explainability of reasoning processes.

Our experimental results underscore the efficacy of \method{} across both drug selection and reasoning generation tasks. For drug selection, \method{} consistently outperformed baseline models, achieving the highest F1 scores on both the xDrugRecommendation and MIMIC-III datasets. In reasoning generation, \method{} demonstrated superior ROUGE scores, highlighting its ability to produce high-quality, interpretable rationales for drug recommendations. Importantly, the inclusion of enriched background information significantly boosted the performance of all evaluated models, further validating the value of integrating domain-specific knowledge into language modeling frameworks.

This work emphasizes the potential of combining graph-based structured knowledge with text-based contextual data to enhance precision and interpretability in drug discovery. By bridging advanced language modeling techniques with biomedical data, \method{} provides a robust foundation for more effective and explainable AI-driven drug discovery pipelines.

In future work, we aim to extend the scalability of \method{} to larger and more diverse datasets, exploring its potential applications in other critical biomedical tasks, such as drug repurposing and adverse effect prediction. These advancements will further contribute to the integration of AI in accelerating therapeutic innovation and improving patient outcomes.


\bigskip

\bibliographystyle{ACM-Reference-Format}
\bibliography{reference}

\end{document}